%% file: main.tex
\documentclass[conf]{new-aiaa}
\usepackage[utf8]{inputenc}

\usepackage{subfiles}
\usepackage{subfigure,setspace}
\usepackage{graphicx}
\usepackage{amsmath}
\usepackage[version=4]{mhchem}
\usepackage{siunitx}
\usepackage{enumitem}
\usepackage{booktabs}

\usepackage{longtable, tabularx, ragged2e}
\newcolumntype{C}{>{\Centering\arraybackslash}X} % centered "X" column

\newcolumntype{M}[1]{>{\centering\arraybackslash}m{#1}} 

\usepackage[flushleft]{threeparttable}
\title{Centralized Copy-Paste: Enhanced Data Augmentation Strategy for Wildland Fire Semantic Segmentation}
\author{Joon Tai Kim\footnote{Ph.D. Student, Mechanical and Aerospace Engineering, AIAA Student Member.}}
\affil{Ohio State University, Columbus, OH 43210}

\author{Tianle Chen\footnote{Ph.D. Student, Department of Computer Science.}}
\affil{Boston University, Boston, MA 02215}

\author{
Ziyu Dong\footnote{Ph.D. Student, School of Environment and Natural Resources.}, 
Nishanth Kunchala\footnote{Undergraduate Student, Mechanical and Aerospace Engineering, AIAA Student Member.} and
Alexander Guller\footnote{Undergraduate Student, Computer Science and Engineering.}
}
\affil{Ohio State University, Columbus, OH 43210}

\author{Daniel Ospina Acero\footnote{Faculty Member, Department of Electronics and Telecommunications Engineering.}}
\affil{Universidad de Antioquia, Medell\'in, Antioquia, 050010, Colombia}

\author{
Roger Williams\footnote{Associate Professor, School of Environment and Natural Resources.} and
Mrinal Kumar\footnote{Professor, Elizabeth Martin Tinkham Endowed Professor of Aeronautical and Astronautical Engineering, AIAA Associate Fellow.}
}
\affil{Ohio State University, Columbus, OH 43210}

\begin{document}
\maketitle

% Note: We will need to revise and update abstract
\begin{abstract}
Collecting and annotating images for training segmentation models is often cost prohibitive, and in wildland fire science this challenge is intensified by the limited availability of reliable public datasets with labeled ground truth. This paper presents the Centralized Copy-Paste Data Augmentation (CCPDA) method, designed to support the training of deep-learning multiclass segmentation models, with particular emphasis on improving segmentation performance for the fire class. CCPDA consists of three main steps: (i) identifying fire clusters in the source image, (ii) applying a centralization technique to isolate the core fire region, and (iii) pasting the refined fire clusters onto a target image. This method increases dataset diversity while preserving the essential characteristics of the fire class. The effectiveness of this augmentation technique is evaluated through numerical analysis and comparisons with alternative augmentation strategies using a weighted-sum multi-objective optimization approach. The results show improved performance metrics for the fire class, which holds greater operational importance than other classes (fuel, ash, or background). Overall, the numerical assessment validates the efficacy of the CCPDA method in alleviating the difficulties associated with small, manually labeled training datasets. It also illustrates that CCPDA outperforms other augmentation strategies in the application scenario considered, particularly in enhancing fire class segmentation performance.
\end{abstract}

% Note: We will need to update the Nomenclature Section 
\section{Nomenclature}
{\renewcommand\arraystretch{1.0}
\noindent\begin{longtable*}{@{}l @{\quad=\quad} l@{}}
$I$                   & Original pixel intensity value. \\
$I'$                  & Adjusted pixel intensity after contrast scaling. \\
$\alpha$              & Contrast scaling factor. \\
$\beta$               & Brightness offset term. \\
$\mathcal{D}$         & Original dataset composed of RGB images and corresponding segmentation masks, i.e., $\mathcal{D} = \{(I_i, M_i)\}_{i=1}^{n}$. \\
$I_i$                 & The $i$-th original RGB image in the dataset. \\
$M_i$                 & The segmentation mask corresponding to image $I_i$. \\
$s_{i,j}$             & The $j$-th fire segment (connected component) in image $I_i$. \\
$S_i$                 & The set of all fire segments in image $I_i$, i.e., $S_i = \{s_{i,j}\}$. \\
$K_5$                 & A square structuring element (kernel) of size 5 pixels used for dilation. \\
$\oplus$              & Morphological dilation operator. \\
$s'_{i,j}$            & The dilated fire segment obtained by applying $\oplus K_5$ to $s_{i,j}$. \\
$R_{\theta}(\cdot)$   & Rotation transformation applied to a segment by an angle $\theta$. \\
$\theta$              & Rotation angle sampled from the uniform distribution $\mathcal{U}(0^\circ, 360^\circ)$. \\
$s''_{i,j}$           & Rotated and dilated fire segment. \\
$S'_i$                & Filtered set of rotated-dilated fire segments with area $\geq 100$ pixels. \\
$n$                   & Total number of original images in the dataset. \\
$r$                   & Number of random seeds used to repeat augmentation, generating multiple samples per pair. \\
$(I'_k, M'_k)$        & Augmented image-mask pair produced by the standard Copy-Paste method. \\
$\mathcal{D}'$        & Augmented dataset generated using the standard Copy-Paste method. \\
$\ominus$             & Morphological erosion operator. \\
$K_x$                 & Square erosion kernel of size $x$ pixels. \\
$\hat{s}_{i,j}$       & Eroded core of fire segment $s_{i,j}$. \\
$\hat{S}_i$           & Set of filtered eroded fire segments from image $i$, used in CCPDA. \\
$(I''_k, M''_k)$      & Augmented image-mask pair created by pasting eroded core segments (CCPDA method). \\
$\mathcal{D}''$       & Augmented dataset generated using the CCPDA method. \\
$\text{Area}(\cdot)$  & Function that computes the number of pixels in a binary segment. \\
$\text{IoU}$          & Intersection over Union. \\
$\text{FNR}$          & False Negative Rate. \\
$F(x)$                & Final evaluation score computed using the weighted sum-based multi-objective optimization method. \\
$w_i$                 & Weight assigned to the $i$-th evaluation metric in the weighted-sum scoring method. \\

\end{longtable*}}

\subfile{sections/01-introduction.tex}
\subfile{sections/02-methodology.tex}

\subfile{sections/03-results.tex}
\subfile{sections/04-conclusion.tex}

%\section*{Appendix}

\section*{Acknowledgments}
The authors acknowledge support for this work from the National Science Foundation award number 2132798 under the NRI 3.0: Innovations in Integration of Robotics Program.

\bibliography{sample}

\end{document}

%% file: sections/01-introduction.tex
\section{Introduction} \label{Introduction}

\subsection{Wildland Fire} \label{Introduction: Wild_Fire}

Wildland fire plays a vital role in maintaining species diversity and ecological balance in fire-dependent ecosystems~\cite{sparks1998effects}. However, the risk of wildfires in the United States has increased rapidly due to increasingly extreme fire behavior, resulting in substantial loss of life and property~\cite{radeloff2023rising, wang2021economic}. This escalation is driven by climate change, abnormal weather patterns, and anthropogenic influences~\cite{brown2023climate}. Given these trends, early fire detection and risk assessment in high fire-risk regions are becoming increasingly important~\cite{harkat2023fire}.

Historical fire regimes have varied widely due to land-use shifts, alterations in vegetation composition, and climate-warming trends~\cite{brown2023climate, abram2021connections, fei2007evidence}. These altered regimes have driven significant changes in the frequency, extent, and severity of wildfires~\cite{montoya2024assessing}. Evidence indicates that compared to the previous two decades, wildfires in the 2000s were up to four times greater, three times more frequent, and more geographically widespread in the United States~\cite{iglesias2022us}. In the eastern U.S., a growing share of large wildfires is caused by human ignition and tends to occur earlier in the year, signaling an expanding fire season~\cite{nagy2018human}. Additionally, eastern hardwood ecosystems have changed from fire-adapted species, such as oaks, to more fire-sensitive, shade-tolerant species such as maples and American beeches~\cite{nowacki2008demise}. These vegetation changes, combined with differences in species flammability, introduce additional uncertainty in modeling future fire behavior and assessing wildfire risk. Therefore, the ability to detect and contain fires at an early stage is critical to minimize ecological and economic losses~\cite{bouguettaya2022review}.

Unmanned aerial systems (UASs) have become effective tools for real-time wildfire surveillance, enabling early detection and monitoring of active fires~\cite{duangsuwan2023accuracy, kim2015real}. A common application of drone technology in wildfire management is the collection of fine-scale, temporally continuous imagery and video data to support firefighting operations~\cite{penglase2023new}. For example, Ford et al. (2024) deployed a UAS platform equipped with an infrared sensor to capture perimeter images in real-time to estimate the spread of the fire \cite{ford2024wildland}. The integration of thermal sensors with drones has also been shown to be effective in monitoring fire temperature and detecting spot fires, demonstrating the viability of this approach for wildland fire detection~\cite{momeni2024collaboration}. In addition, color-based methods using red-green-blue (RGB) cameras have been applied to classify pixels as fire or non-fire~\cite{dang2019aerial}. When coupled with texture analysis, RGB imagery can assist in identifying irregular flame and smoke patterns~\cite{hossain2020forest}. Despite the growing availability of UAS-derived wildland fire imagery, effectively leveraging these datasets for fire behavior analysis and accurate segmentation remains a significant challenge.

Various vision-based fire detection techniques have been developed; however, the most effective approach for accurately classifying and segmenting fire imagery, particularly in UAV-based monitoring, remains unclear~\cite{dang2019aerial, jiao2018forest, yuan2017aerial}. Manual annotation of fire and non-fire regions is commonly used to label fire images~\cite{chen2022wildland}, but this process is prone to bias and uncertainty, especially due to resolution constraints and smoke occlusion. Although advanced algorithms have been introduced to reduce missed fire detections~\cite{li2020image}, their performance has not been fully validated in realistic forest or prairie fire scenarios. In addition, segmentation followed by classification has been used to delineate burned areas in large-scale satellite imagery; however, these methods have not yet been adapted for fine-scale, localized fire scenes~\cite{singh2024beyond}. Therefore, there remains a critical need for accessible and robust segmentation techniques specifically designed for wildland fire contexts, particularly those suited for practical deployment in forest and prairie fire monitoring and management.

\subsection{Deep Learning Models and Data Constraints} \label{Introduction: DL Models and Data Constraints}

Deep learning models, such as U-Net~\cite{ronneberger2015u}, have demonstrated robust performance in image segmentation tasks. However, their success critically depends on access to large, diverse, and well-annotated datasets. In many domains, including the field of medical imaging~\cite{ren2022object}, insufficient labeled data significantly hinders model generalization and leads to poor segmentation results. This issue becomes especially problematic in safety-critical applications, such as wildland fire image segmentation, where misclassification can compromise detection of fire and the effectiveness of response decisions.

Unfortunately, obtaining labeled data in the context of wildland fire imagery is both labor-intensive and time-consuming~\cite{paton2019automating}. Annotating ash, fire, vegetation, and background pixels across diverse environmental conditions---including smoke occlusion, inconsistent lighting, and deformable flame structures---presents a significant challenge. Additionally, public datasets such as Flame~\cite{shamsoshoara2020dataset} and Flame 2~\cite{hopkins2022dataset} are limited in scope, primarily providing only binary segmentation: fire versus non-fire annotations. Although suitable for basic research, these binary datasets lack the detail required for real-world wildfire management tasks. In practice, identifying fire alone is often insufficient; accurate segmentation of fuel sources (vegetation) is essential for predicting fire spread and understanding burn dynamics, while mapping ash regions helps delineate already-burned areas and assess wildfire containment.

For effective fire spread prediction and scene understanding, it is essential to advance beyond binary classification. Distinguishing not only fire, but also vegetation (fuel source), ash (burned areas), and background is necessary for accurate modeling and operational decision support~\cite{hiers2020prescribed}. However, the creation of such fully labeled datasets remains a bottleneck, limiting the development and deployment of robust image segmentation models in this domain.

\subsection{Data Augmentation} \label{subsec:related_work_data_augment}

Data augmentation has become a cornerstone in computer vision, particularly when training data is limited or imbalanced. The field has progressed from simple geometric and photometric transformations to increasingly sophisticated, task-specific augmentation pipelines. Traditional approaches relied on encoding invariances to common transformations, such as random cropping and horizontal flipping~\cite{lecun1998gradient, krizhevsky2012imagenet}, to artificially expand dataset diversity. Extending these approaches, policy-based methods like AutoAugment~\cite{cubuk2019autoaugment} and RandAugment~\cite{cubuk2020randaugment} leveraged reinforcement learning and the randomized search approach to discover augmentation strategies that optimize downstream performance. Significant advancement came with mixing-based methods such as CutMix~\cite{zhang2019bag}, which generate synthetic examples by blending image regions and their corresponding labels. These techniques improved generalization and facilitated spatially-aware learning. For dense prediction tasks such as segmentation, object-aware augmentations became essential. Approaches like ``Cut, Paste and Learn''~\cite{dwibedi2017cut} and InstaBoost~\cite{fang2019instaboost} synthesized novel scenes by copying object instances and pasting them onto new contexts with appropriate geometric transformations. Copy-Paste augmentation~\cite{ghiasi2021simple} further demonstrated strong performance gains in instance segmentation by leveraging object compositionality across training samples.

Despite the success of these techniques in general vision tasks, data augmentation remains largely unexplored in fire-related visual analysis. Most existing studies rely on conventional color-space based augmentations, such as random chromatic distortion in HSV space~\cite{9067475}, which may be insufficient to model the complex and irregular visual properties of fire. The complex visual characteristics of fire present unique challenges for segmentation models, especially in safety-critical domains like wildland fire analysis, where accuracy in boundary detection and object class differentiation is crucial. Consequently, there is a pressing need for augmentation methods specifically designed to address the distinctive visual properties of fire imagery.

\subsection{Objectives and Contributions} \label{Introduction: Objective}

\begin{figure}[!htb]
    \centering
    \includegraphics[width=0.95\textwidth]{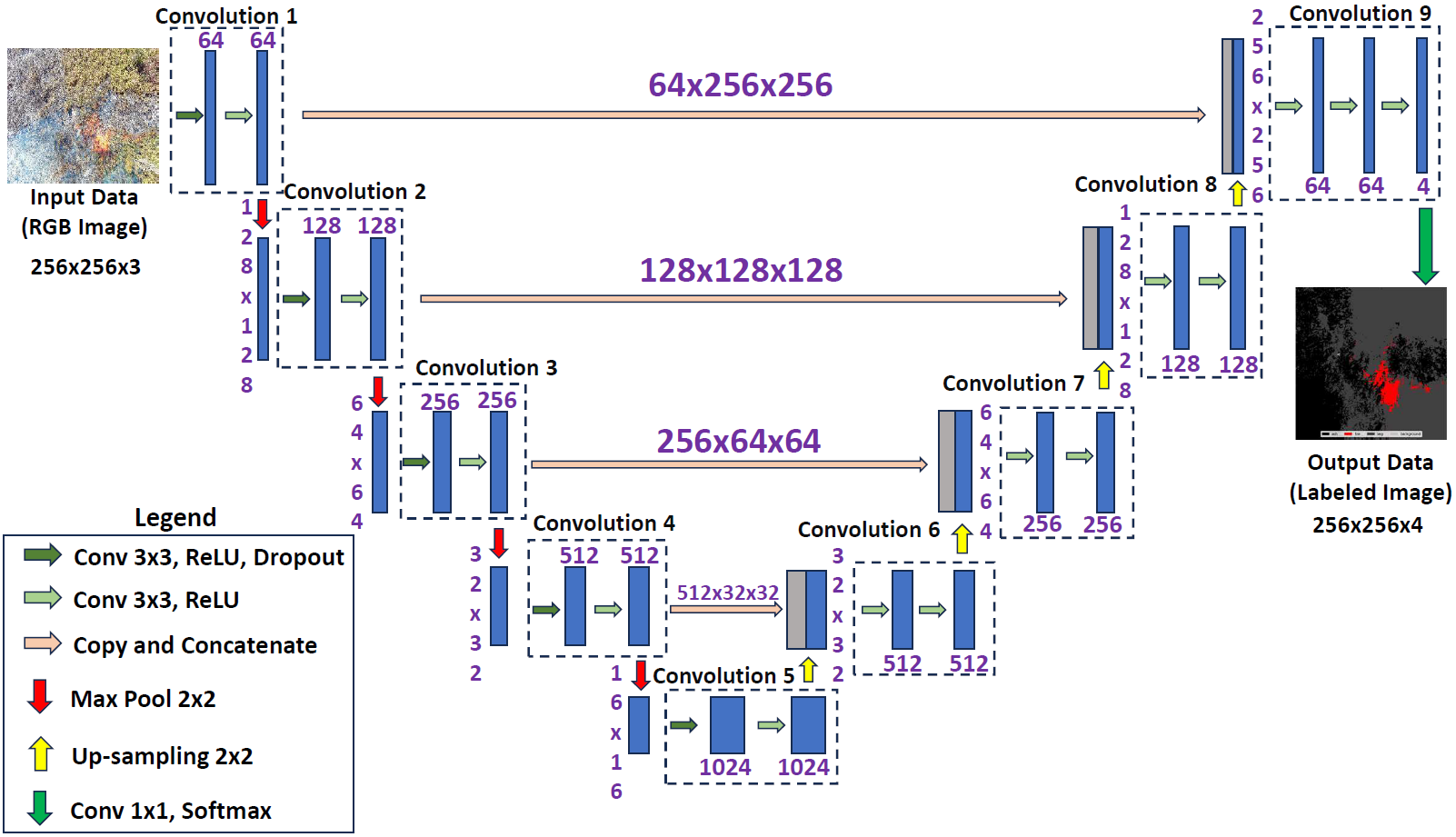}
    \caption{The U-Net architecture used as the base-line model.}
    \label{fig:unet_structure}
\end{figure}

The main objective of this study is to improve the semantic segmentation of wildland fire imagery under limited labeled data conditions, with an emphasis on reducing missed detections of the fire class. To address this, we introduce a novel data augmentation methodology called Centralized Copy-Paste Data Augmentation (CCPDA), which enhances segmentation performance by pasting the core regions of fire clusters---where labels are most reliable---onto new wildland fire images, while omitting ambiguous boundary pixels prone to misclassification. This strategy is integrated into a U-Net segmentation framework (see Fig.~\ref{fig:unet_structure}), a widely used architecture for pixel-wise classification tasks. To ensure that training is guided by metrics relevant to real-world applications, we employ a weighted-sum multi-objective optimization (MOO) strategy that minimizes false negatives in the fire class while maintaining strong performance on vegetation and overall segmentation quality. For this study, we utilized our multiclass wildfire dataset BURN 1~\cite{BURN01} within the U-Net–based semantic segmentation methodology to evaluate the effectiveness of the proposed CCPDA method. The main contributions of this work are summarized as follows:
\begin{enumerate}
    \item We introduce the Centralized Copy-Paste Data Augmentation (CCPDA) method to improve the training of multiclass deep learning segmentation models in wildland fire imagery. CCPDA is applied to augment UAV-captured RGB images by selectively pasting the interior regions of fire clusters that are less susceptible to labeling errors onto new images.
    \item We develop and employ a weighted-sum multi-objective optimization (MOO) framework to evaluate and rank data augmentation methods based on performance metrics that are operationally critical for wildfire response. In particular, the framework emphasizes minimizing false negatives in the fire class, as missed fire detections can severely hinder early response and decision-making for wildfire management, while also accounting for contributions from other segmentation classes.
    \item We construct a novel four-class wildfire image dataset, BURN 1~\cite{BURN01}, consisting of ash, fire, vegetation, and background classes. The dataset includes 20 manually annotated UAV images collected during a prescribed burn on a grassland prairie in central Ohio, providing high-quality labels for training and evaluating segmentation models.
\end{enumerate}

% U-Net Figure 

%% file: sections/02-methodology.tex
\section{Methodology} \label{Methodology}
% Introduction Paragraph 
In this section, we detail the process of generating the CCPDA method and validating its effectiveness through comparative analysis with other augmentation strategies. Our approach is organized as follows: 1) data collection from real wildfire scenes; 2) data preprocessing, which includes dehazing and ground truth labeling; 3) data augmentation using color-based methods such as brightness and contrast adjustments, image rotations, and Copy-Paste techniques; 4) evaluation of these augmentation methods through a weighted sum-based MOO process that combines Intersection over Union (IoU) and False Negative Rate (FNR) scores, with particular emphasis on the fire class; and 5) tuning of U-Net model hyperparameters---learning rate, batch size, and dropout rate---via grid search, using the weighted sum-based score on the test dataset to select the most effective configuration.

\subsection{Data Collection} \label{Methodology: Data_Collect}
The dataset for this study was collected during a prescribed burn in a section of the Larry Yoder Experimental Grassland Prairie at the Marion Campus of The Ohio State University (OSU). The date of the prescribed burn was October 8, 2022. The images were captured using a DJI Mavic 2 drone equipped with an RGB camera, flying at altitudes between 4 and 128 meters, with the camera oriented in a nadir position for bird's-eye view. The original RGB images were taken under fair lighting conditions on a sunny autumn day between 1:00 and 2:30 PM. The Vantage Vue Wireless Integrated Sensor Suite Weather Station recorded the weather data every minute during that interval, with summary statistics as detailed in Table \ref{table:weather_conditions}.

% Weather Stat Table 
\begin{table}[!htb]
    \centering
    \caption{Weather conditions during the prescribed burn.}
\begin{tabular}{|c|c|c|c|}
    \hline
    \textbf{} & \textbf{Temperature (°C)} & \textbf{Humidity (\%)} & \textbf{Wind Velocity$^{*}$ (km/h)} \\ \hline
    \textbf{Mean} & 12.49 %4923 
    & 42.91 %121 
    & 0.59 
     
    \\ \hline
    \textbf{Std. Deviation} & 0.59 
    %5912 
    & 1.71 
    %074 
    & 1.04 
    %0.9751 
    \\ \hline
  \multicolumn{4}{l}{\footnotesize $^{*}$These numerical values simply indicate calm conditions and should not be interpreted as a normal distribution.}\\[-0.5ex]
    \end{tabular}%
    \label{table:weather_conditions}
\end{table}

\subsection{Data Preprocessing} \label{Methodology: Data_PreProcess}

\subsubsection{Smoke Dehazing} \label{Methodology dataPreProcess: Smoke_Dehaze}
% Dehaze Flowchart Figure 
\begin{figure}[!htb]
    \centering
    \includegraphics[width=0.6\textwidth]{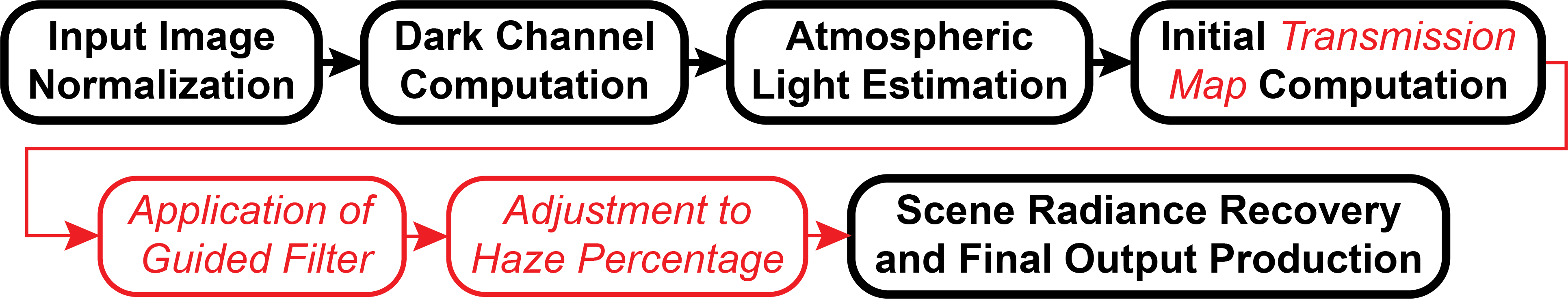}
    \caption{Flowchart for smoke dehazing method.}
    \label{fig:dehaze_flowchart}
\end{figure}

% Dehaze Output Images 
\begin{figure}[!htb]
    \begin{center}
        \subfigure[Original Hazy Image]
        {\resizebox{3.0in}{!}{\includegraphics{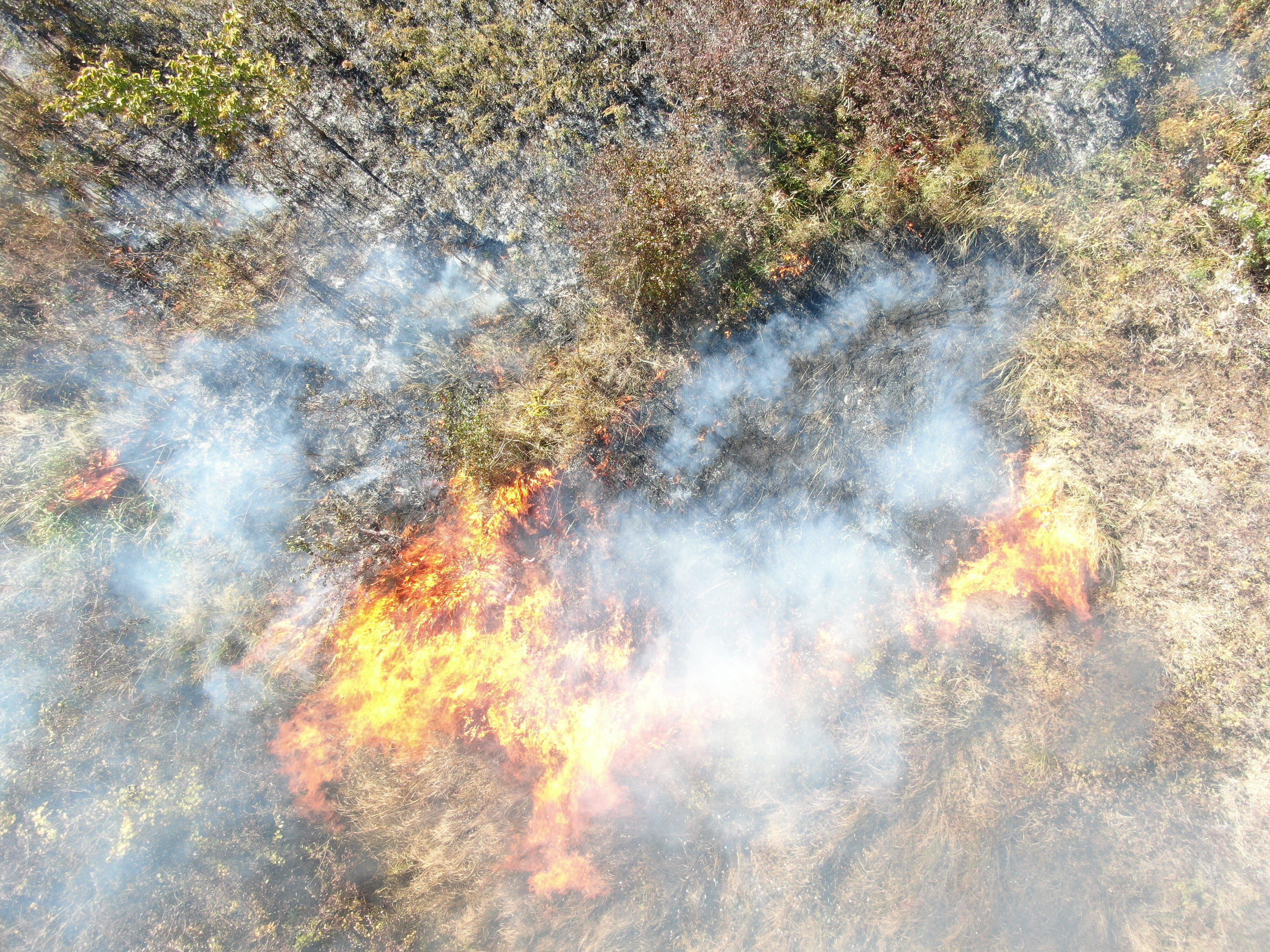}} \label{fig:dehazing_a}}
        \subfigure[Dehazed Image (95\% Reduction)]
        {\resizebox{3.0in}{!}{\includegraphics{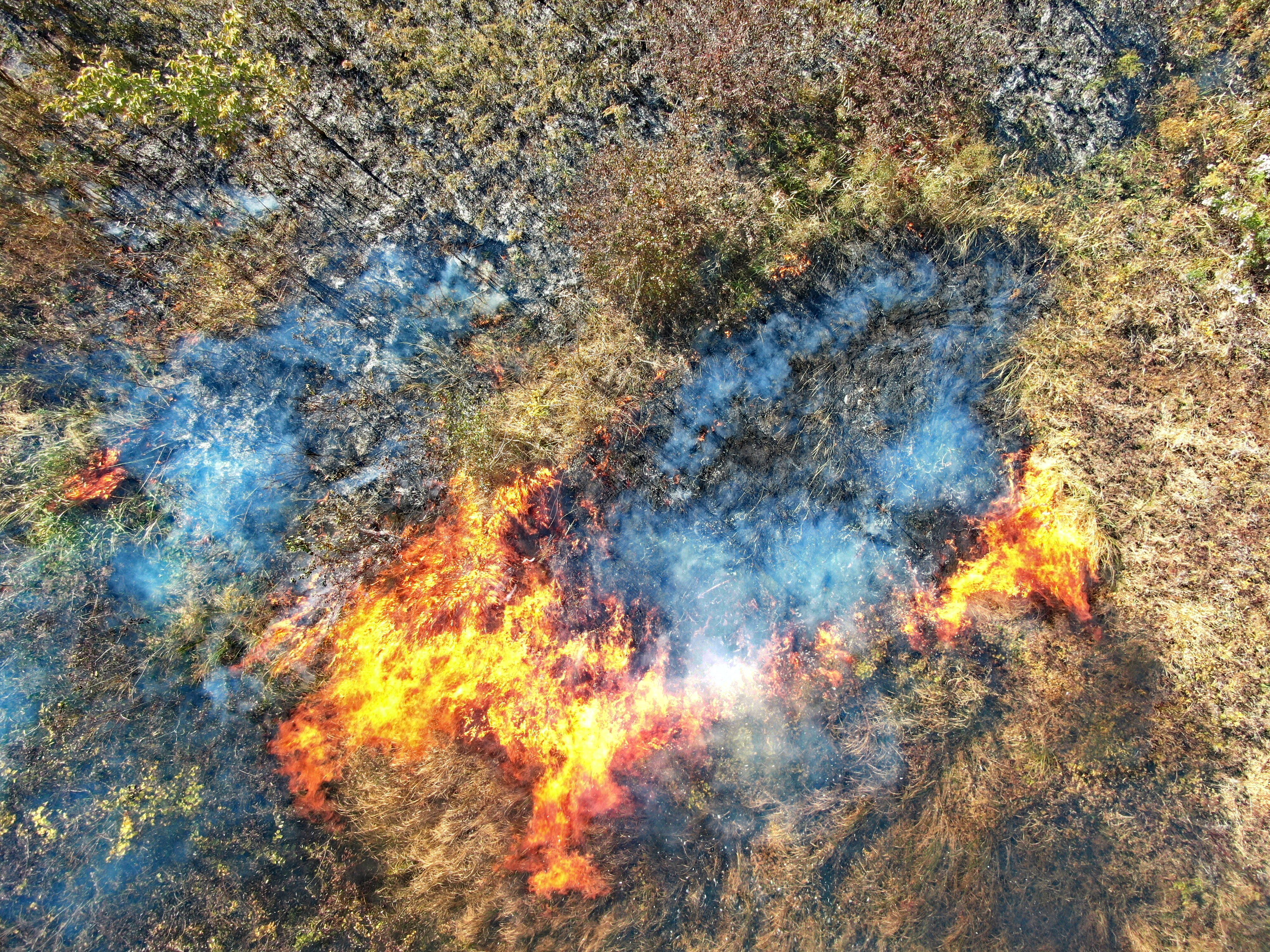}} \label{fig:dehazing_b}}
    \end{center}
    \caption{A comparison of the original hazy RGB image (a) and the dehazed RGB image (b) achieved through a 95\% haze-reduction procedure.}
    \label{fig:dehaze_compare}
\end{figure}

First, 20 images characterized by varying degrees of smoke and fire clusters were selected for our dataset. To minimize human labeling errors during ground truth annotation, we applied a dehazing algorithm based on the method proposed by He et al.~\cite{he2011dehaze}, which utilizes the dark channel prior: a statistical approach to estimate haze. However, to refine the transmission map, we elected to use the guided filter method presented in \cite{he2013guided} and \cite{pang2011improved-guided}, instead of the soft matting process, to reduce computational cost. Fig. \ref{fig:dehaze_flowchart} illustrates the flowchart of the dehazing pipeline. Finally, Fig. \ref{fig:dehazing_a} and Fig. \ref{fig:dehazing_b} present a visual comparison between the original hazy image and the result of the dehazing process, respectively, demonstrating the effectiveness of the 95\% haze-reduction procedure.

\subsubsection{Ground Truth Annotation} \label{Methodology dataPreProcess: GT_Annotate}
Ground truth labels are essential for training deep learning-based segmentation models such as U-Net. In this work, labels were generated using the ImageJ software to differentiate four classes: ash, fire, vegetation, and background. Specifically, we manually annotated several sample pixels strongly associated with each class, selected based on their distinct color characteristics. The Labkit plug-in’s default random forest-based pixel classification tool was then used to generate preliminary segmentation labels for the entire image \cite{arzt2022labkit}. These results were subsequently reviewed for misclassified pixels produced by the software, which we manually corrected. Although the process is somewhat iterative, it offers a practical and user-friendly approach to ground truth labeling for RGB images.

\subsection{Data Augmentation} \label{Methodology: Data_Aug}
% Intro part for Data Augmentation 
In this section, we describe the data augmentation methods considered in this work: rotation-based augmentation, color-based augmentation (including brightness and contrast scaling), and the Copy-Paste augmentation technique. These methods are experimentally compared in Sections~\ref{Results: Study01} and~\ref{Results: Study02}, first evaluating the efficiency of the Copy-Paste method against the baseline non-augmentation approach, the rotation-based augmentation, and the color-based techniques. Next, the analysis of the Copy-Paste method is extended to determine whether fire class segmentation performance can be improved by mitigating the impact of human labeling errors. To do this, we propose the CCPDA method, which refines the copied fire clusters by excluding edge regions that are more prone to inter-class mislabeling. Finally, varying degrees of the CCPDA method--- defined by how much is cropped from the original fire cluster ---are compared to evaluate the effectiveness of this approach in enhancing segmentation performance. 
It is important to note that, following each augmentation process described below, all original images were downscaled from 4000×3000 to 256×256 prior to input into the U-Net segmentation model, to comply with computational constraints. The dataset was initially split into training and testing sets in a 5:5 ratio, with the training set further divided into training and validation subsets using an 8:2 split. This resulted in a final distribution of 8:2:10 for training, validation, and testing, respectively. The test set, serving as a control group, was left unaugmented to evaluate model performance on non-augmented, unaltered data.

\subsubsection{Rotation Based Method} \label{Methodology dataAug: Rotation_Method}
For the rotation-based augmentation process, we used a 15-degree rotation increment to produce augmented images. This angle was chosen to ensure consistency with the dataset size produced by the Copy-Paste approach. Starting with eight original training images, we generated a total of 192 augmented images using this rotation scheme.
A key challenge with rotation-based augmentation is the potential loss of image content when parts of the image rotate out of frame. Additionally, the resulting out-of-frame areas are filled with black borders (RGB value [0, 0, 0]), which in our multi-class dataset are strongly associated with the ash class. This may lead the model to misinterpret border artifacts as ash, introducing noise and degrading model performance. To address this, we preemptively scaled the original images by a factor of 1.66; the minimum required to avoid black padding of edges during rotation. The scaled images were then rotated from 5 degrees to 360 degrees in 15-degree increments. Beginning at 5 degrees (instead of 0) prevents duplicate inclusion of the original, unrotated images already reserved for the test set. Finally, each rotated image was cropped back to its original dimensions to ensure a centered, uniform input.

\subsubsection{Brightness Scaling Method} \label{Methodology dataAug: Brightness_Method}
For the brightness-based augmentation process, we applied scaling factors to the brightness channel of each image in HSV color space. Each RGB image was first converted to HSV, and the Value (V) channel was multiplied by 24 different scaling factors ranging from 1.00 (original brightness) to 2.15, in increments of 0.05. To ensure that all pixel values remained within valid bounds, the adjusted intensities were clipped to the 8-bit range [0, 255] after scaling. The modified image was then converted back to RGB color space and saved for training. Starting with eight original training images, this procedure produced a total of 192 augmented images, matching the dataset size used in the rotation-based and Copy-Paste methods. Importantly, corresponding ground truth masks were duplicated without modification, as brightness adjustments do not affect image geometry or class boundaries.

\subsubsection{Contrast Scaling Method} \label{Methodology dataAug: Contrast_Method}
For the contrast-based augmentation process, we applied 24 different scaling factors ranging from 0.50 to 1.65 in increments of 0.05. For each pixel, the new intensity value was computed as:  
\begin{equation}
    I' = I \cdot \alpha + \beta \ ,
\end{equation} 
where \( I \) is the original intensity, \( \alpha \) is the contrast scaling factor, and \( \beta \) is a brightness offset set to zero. Values of \( \alpha > 1 \) increase contrast, values between 0 and 1 reduce contrast, and \( \alpha = 1 \) preserves the original contrast.

Starting with eight original training images, this process produced a total of 192 augmented images, consistent with the dataset sizes used in other augmentation methods. For the same reason as mentioned in Section~\ref{Methodology dataAug: Brightness_Method}, the ground truth masks for contrast scaling were duplicated without modification.

\subsubsection{Standard Copy-Paste Method} \label{Methodology dataAug: Std_CP_Method}
% Note: Tianle
To address the limited sample diversity in the original wildfire dataset and the highly variable morphology of flame structures, we adopt a class-specific data augmentation technique inspired by the Copy-Paste strategy introduced in~\cite{google-Copy-Paste}. This approach improves generalization by artificially increasing the diversity of fire segment appearances and improving class balance within the training dataset. The core idea behind this method is to identify individual fire clusters in the source image and paste them onto different target images, while applying transformations such as dilation and random rotation to increase visual variability. This approach preserves the shape and internal properties of the fire clusters by copying complete segments and integrating them with minimal visual artifacts. 

The original dataset is denoted by:
\begin{equation}
    \mathcal{D} = \{(I_i, M_i)\}_{i=1}^{n} \ ,
\end{equation}
where $I_i$ is the $i$-th RGB image, and $M_i$ is its corresponding pixel-wise segmentation mask. For each image $I_i$, we extract the connected fire components from $M_i$ by identifying all the disjoint pixel clusters corresponding to the fire class:
\begin{equation}
    S_i = \left\{ s_{i,j} \,\middle|\, s_{i,j} \subseteq M_i,\ \text{class}(s_{i,j}) = \text{fire},\ j=1,2,\dots,N_i \right\} \ ,
\end{equation}
where $s_{i,j}$ denotes the $j$-th fire segment in image $i$, and $N_i$ is the number of fire segments in $M_i$. To incorporate additional boundary pixels around each fire cluster and reduce blending artifacts, each segment is dilated using morphological operations. Specifically, we apply a dilation operator with a 5-pixel square structuring element $K_5$:
\begin{equation}
    s'_{i,j} = s_{i,j} \oplus K_5 \ ,
\end{equation}
where $\oplus$ denotes dilation, which expands the boundary of the binary segment $s_{i,j}$ to include pixels adjacent to the background or in the class. This expansion helps the pasted region blend more naturally into the target background. To further increase diversity, we apply a random in-plane rotation to each dilated segment:
\begin{equation}
    s''_{i,j} = R_{\theta}(s'_{i,j}), \quad \theta \sim \mathcal{U}(0^\circ, 360^\circ) \ ,
\end{equation}
where $R_\theta(\cdot)$ denotes a rotation operator by angle $\theta$, and the rotation angle is sampled from a uniform distribution over $[0^\circ, 360^\circ]$. This step allows the model to observe the same fire shape under various orientations, which is particularly important for aerial imagery, where the viewpoint can vary. After rotation, we apply a size threshold to discard small or fragmented components that may be noise or irrelevant features:
\begin{equation}
    S'_i = \left\{ s''_{i,j} \,\middle|\, \text{Area}(s''_{i,j}) \geq 100 \text{ pixels} \right\} \ .
\end{equation}

The Copy-Paste augmentation process then constructs a new dataset by pasting segments from $S'_i$ onto other images $I_j$ (with $i \neq j$), and updating the corresponding label masks accordingly. To generate sufficient data diversity, we repeat this process $r$ times using different random seeds to vary the placement of segments:
\begin{equation}
    \mathcal{D}' = \left\{ (I'_k, M'_k) \right\}_{k=1}^{n^2r} \ ,
\end{equation}
where each augmented image-label pair $(I'_k, M'_k)$ is produced by sampling a source image $I_i$ and a target image $I_j$ from $\mathcal{D}$, and blending a randomly selected subset of $S'_i$ into $I_j$ at a valid position. This method significantly increases the size and variability of the dataset while preserving the meaningful visual and structural properties of the pasted fire clusters, improving the model’s ability to generalize to unseen fire configurations during inference. Fig.~\ref{fig:ccpda_pipeline}, specifically the bottom portion, illustrates the Standard Copy-Paste procedure in visual detail.

\subsubsection{Centralized Copy-Paste Method} \label{Methodology dataAug: CCPDA}
% Note: Tianle
% CCPDA Figures 
\begin{figure}[!htb]
    \begin{center}
        \subfigure[Copy-Paste Augmentation Pipeline]
        {\resizebox{3.4in}{!}{\includegraphics{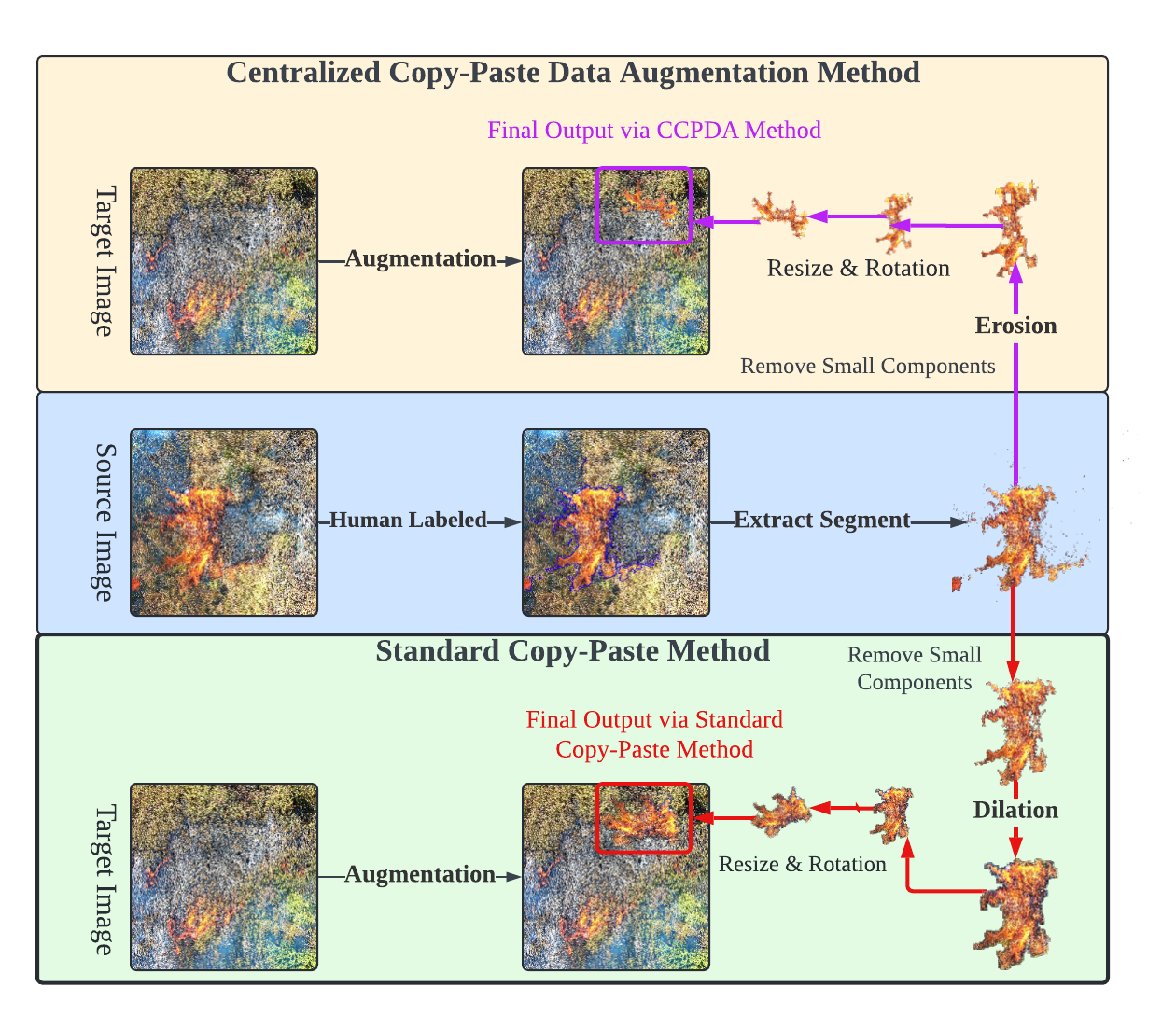}} \label{fig:ccpda_pipeline}}
        \hspace{0.2in}
        \subfigure[CCPDA Applied to a Single Fire Cluster]
        {\raisebox{0.8in}{\resizebox{2.6in}{!}{\includegraphics{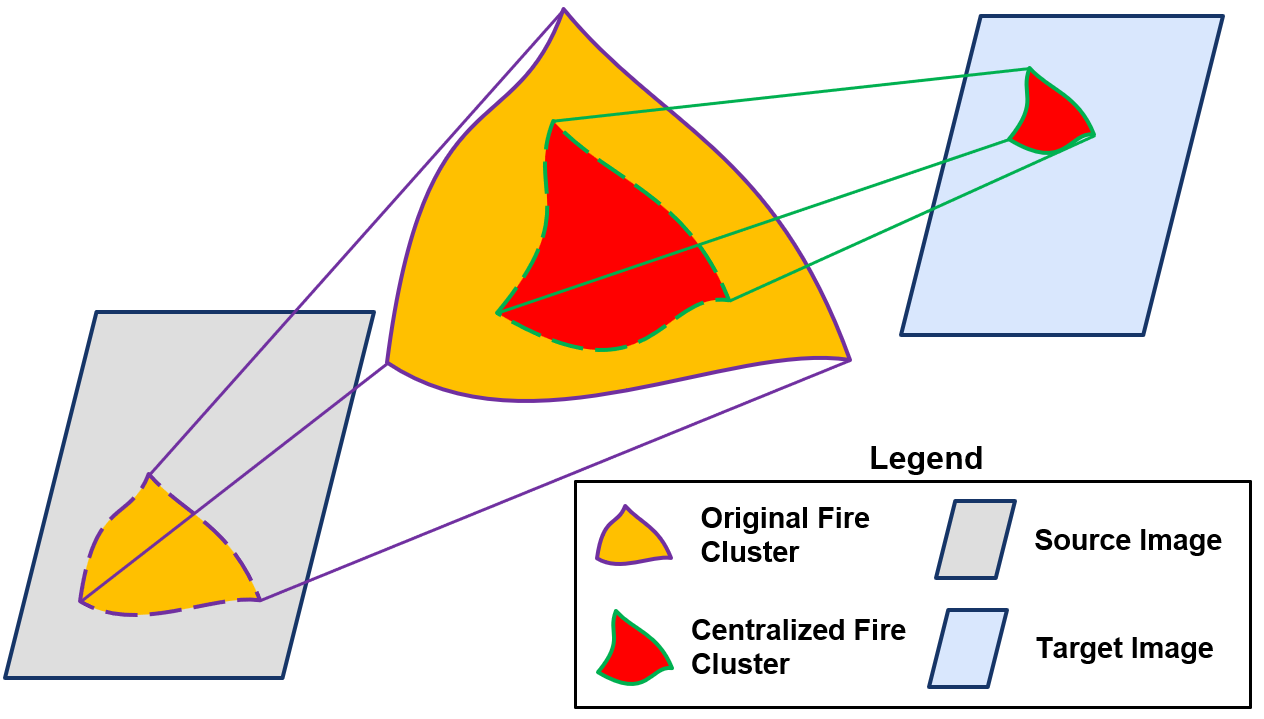}}} \label{fig:centralized_diagram}}
    \end{center}
    \caption{(a) Copy-Paste augmentation pipeline illustrating both the Standard (bottom) and Centralized (top) methods. (b) Visual representation of the centralized copy-paste operation on a single fire cluster, where only the centralized core (red) of the fire cluster (yellow) is copied from the source image (gray) to the target image (light blue).}
    \label{fig:ccpda_figures}
\end{figure}

Although the standard Copy-Paste method improves fire diversity and generalization, it also propagates errors associated with weak or uncertain labels, especially along the boundaries of fire clusters. These boundary fire pixels are often mislabeled due to occlusions (e.g., smoke) or visual similarity to adjacent vegetation, which can result in misleading supervision during training. To mitigate this, we introduce a refinement to the standard method, named the Centralized Copy-Paste Data Augmentation (CCPDA) method. This approach systematically removes uncertain boundary pixels and focuses on the high-confidence core regions of fire.

The CCPDA method modifies the segment selection step in the standard Copy-Paste pipeline by applying morphological erosion to each fire segment. $s_{i,j}$ is a binary mask of a single fire segment as defined in the previous Section~\ref{Methodology dataAug: Std_CP_Method}. Then, we define the eroded segment as:
\begin{equation}
    \hat{s}_{i,j} = s_{i,j} \ominus K_x \ ,
\end{equation}
where $\ominus$ denotes morphological erosion and $K_x$ is an $x$-pixel square kernel. This operation shrinks the segment by removing $x$ pixels from its boundary in all directions, thus retaining only the central core regions, which are regions most certainly associated with true fire due to inherent fire dynamics. As in the standard method, we apply a minimum area constraint to remove overly small or noisy cores:
\begin{equation}
    \hat{S}_i = \left\{ \hat{s}_{i,j} \,\middle|\, \text{Area}(\hat{s}_{i,j}) \geq x \text{ pixels} \right\} \ .
\end{equation}
These high-confidence segments $\hat{S}_i$ are then used to generate a new augmented dataset by pasting them onto target images as in the original Copy-Paste method:
\begin{equation}
    \mathcal{D}'' = \left\{ (I''_k, M''_k) \right\}_{k=1}^{n^2r} \ .
\end{equation}

In practice, the kernel size $x$ controls the applied erosion size and serves as a tunable parameter for balancing the trade-off between retaining the core fire structures and removing ambiguous borders. By concentrating the augmentation on the most reliable regions of each fire segment, the CCPDA method improves training stability and reduces the risk of incorporating mislabeled pixels. This is particularly valuable in scenarios where fire boundaries are difficult to accurately annotate and where small labeling errors can significantly degrade segmentation performance. Taken together, the standard and centralized Copy-Paste methods form a comprehensive augmentation framework tailored for pixel-wise semantic segmentation of fire imagery. The CCPDA refinement is specifically designed to reduce noise propagation and improve model robustness in the presence of label uncertainty, a common challenge in high-risk environmental monitoring scenarios such as wildfire detection. Fig.~\ref{fig:ccpda_pipeline}, specifically the top portion, illustrates the CCPDA procedure, while Fig.~\ref{fig:centralized_diagram} provides a detailed visual representation of a single centralized fire cluster being copied and pasted onto a target image.

\subsection{Image Segmentation Procedure} \label{Methodology: Img_Seg}
To evaluate the effectiveness of the proposed image augmentation methods, we employ a U-Net architecture for semantic segmentation. Originally developed for biomedical image segmentation \cite{ronneberger2015u}, U-Net has become a widely used framework for pixel-wise classification, particularly in scenarios with limited training data \cite{shah2023wildfire,ribeiro2023burned}. Its U-shaped encoder-decoder structure enables efficient feature extraction and precise localization. A schematic of the U-Net configuration employed in this study is presented in Fig.~\ref{fig:unet_structure}.

\subsection{Evaluation of Augmentation Methods} \label{Methodology: Eval_Aug_Methods}
The weighted sum-based MOO method is used to evaluate model performance across different data augmentation techniques by simultaneously optimizing multiple ``competing'' evaluation metrics \cite{chakraborty2023moo}. To aggregate and rank the scores of each segmentation method, we use the weighted sum method, which calculates a single score as a linear combination of evaluation metrics, assigning specific weights to prioritize the most important metric, while still considering less-important metrics \cite{marler2010weighted}. 

Given the focus on supporting wildfire management decision-making by minimizing missed wildfire detections and improving fire-spread prediction, the following three evaluation metrics were identified as the most important: fire-FNR, vegetation-IoU, and total-IoU. The two key metrics, IoU and FNR, are expressed as 
\begin{equation}
    \notag \text{IoU} = \frac{TP}{TP + FP + FN} \ ; \ \
    \text{FNR} = \frac{FN}{TP + FN} \ .
\end{equation} 
These metrics were selected for specific reasons: minimizing fire-FNR is crucial due to the life-threatening risks of missed detection; vegetation-IoU ensures accurate assessment of fuel sources for real-time fire-spread prediction; and total-IoU provides a comprehensive measure of the model's overall performance. The weighted sum scores are computed as
\begin{equation}
    F(x) = w_1 f_{\text{fire-FNR}}(x) + w_2 f_{\text{veg-IoU}}(x) + w_3 f_{\text{total-IoU}}(x) \ ,
    \label{eq:weighted_score}
\end{equation}
where \( w_i \) are the individual weights for each metric. We chose to use the rank-order centroid (ROC) method \cite{ahn2011roc} for its simplicity and intuitive approach to computing weights based on a ranking scheme. This method calculates weights using the equation
\begin{equation}
    w_i = \frac{1}{n} \sum_{k=i}^{n} \frac{1}{k} \ ,
    \label{eq:roc_weights}
\end{equation}
where \( n \) represents the total number of metrics and \( k \) is the rank position of each metric. Since we are using only 3 metrics, the resulting weights in our case are \( w_1 = 0.611 \), \( w_2 = 0.278 \), and \( w_3 = 0.111 \). It is important to note that minimizing the FNR corresponds to better performance, so we define \( f_{\text{fire-FNR}}(x) = (1 - \text{FNR}) \).

Finally, using the weighted sum-based scoring method, we conducted a numerical comparison to demonstrate that the Copy-Paste method outperforms other data augmentation techniques. Additionally, we performed a second comparative analysis to evaluate the model’s performance when centralized fire clusters generated with varying erosion values are added to target images. These studies are structured as follows:
\begin{itemize}
    \item \textbf{Study 1}: Comparison of augmentation methods: non-augmented, rotation-by-15$^\circ$, and standard Copy-Paste,
    \item \textbf{Study 2}: Comparison of CCPDA method with different erosion values: 0\%, 10\%, 20\%, and 30\%.
\end{itemize}

The purpose of Study 1 is to show that the Copy-Paste augmentation method is superior due to the increased diversity it introduces to the training dataset, which is beneficial for model performance. Study 2 aims to address human labeling errors that occur at the borders of fire clusters, particularly the difficulty in distinguishing between fire and vegetation, where their RGB values can be similar. These errors can negatively impact the model’s segmentation performance, and by removing these prone-to-error edges, we can show that the fire class segmentation performance improves. It is also important to note that during Study 2, the rotation angle and the location \((x, y)\) of the pasted fire segment on the target image were kept constant---specifically at \((0.25, 0.25)\)---to ensure comparability and reduce variability in the comparative analysis, avoiding any randomness that could obfuscate the interpretability of the results.

\subsection{Hyperparameter Tuning}
\label{Methodology: Hp_Tune}
% Note: Nishanth
To identify the optimal model configuration, we conducted a grid search on three key hyperparameters: learning rate, dropout rate, and batch size. The tuning process was implemented through Python, using the TensorFlow library, to automate training, evaluation, and model selection.

\subsubsection{Composite Scoring Function} \label{Hyperparameter Tuning: Comp_Scoring_Func}
Both F(x), the weighted sum-based composite score defined in 
Equation~\ref{eq:weighted_score}, and the associated weights derived via the rank-order centroid (ROC) method in Equation~\ref{eq:roc_weights}, were previously introduced in Section~\ref{Methodology: Eval_Aug_Methods}. As previously emphasized, this formulation prioritizes reducing fire false negative rates, while incorporating vegetation and overall IoU, providing a balanced, yet fire-focused assessment of segmentation performance.

\subsubsection{Search Space} \label{Hyperparameter Tuning: Search_Space}
The search space included the following candidate values:

\begin{itemize}
    \item Learning rate: \{0.01, 0.005, 0.001, 0.0005\}
    \item Dropout rate: \{0.0, 0.1, 0.2, 0.3\}
    \item Batch size: \{4, 8, 16\}
\end{itemize}

This search space resulted in $4 \times 4 \times 3 = 48$ unique model configurations. Each configuration was trained for 80 epochs using the Adam optimizer and categorical cross-entropy loss. Performance metrics were automatically recorded in a CSV file at each iteration. The first model was stored as a baseline, and subsequent models were retained only if their \( F(x) \) score exceeded that of the current best. All non-optimal model checkpoints were discarded to conserve storage and streamline analysis.

\subsubsection{Best Configuration} \label{Hyperparameter Tuning: Best_config}
The top 3 configurations, ranked by $F(x)$, together with the bottom three configurations with the lowest $F(x)$ scores, are shown in Table~\ref{tab:tuning-summary}. It is important to note that the first row indicates the final selected hyperparameter setting used in this study. Furthermore, we can observe that while vegetation IoU and overall IoU remain relatively stable across configurations (with variations within approximately 0.2\%), the false negative rate of the fire class exhibits substantial variability—ranging from as low as 4.25\% to as high as 100.00\%.

\begin{table}[!htb]
\centering
\caption{Selected hyperparameter configurations ranked by $F(x)$ score.}
\label{tab:tuning-summary}
\begin{tabular}{@{}ccccccc@{}}
\toprule
Learning Rate & Dropout Rate & Batch Size & $F(x)$ & Fire FNR [\%] & Veg IoU [\%] & Test IoU [\%] \\
\midrule
\textbf{0.0005} & \textbf{0.3} & \textbf{8}  & \textbf{0.83009} & \textbf{4.25} & \textbf{65.98} & \textbf{55.52} \\
0.001  & 0.2 & 4  & 0.83006 & 5.49 & 68.05 & 57.16 \\
0.0005  & 0.3 & 4  & 0.82856 & 5.51 & 67.45 & 57.40 \\
\multicolumn{7}{c}{$\vdots$} \\
0.005 & 0.0 & 16  & 0.22205 & 100.00 & 61.95 & 44.88 \\
0.005 & 0.2 & 8  & 0.22205 & 100.00& 61.95 & 44.88 \\
0.001 & 0.0 & 4  & 0.22205 & 100.00 & 61.95 & 44.88 \\
\bottomrule
\end{tabular}
\end{table}

This variability suggests that the model’s ability to detect fire pixels is highly sensitive to changes in hyperparameters, particularly dropout and learning rate. Given that fire regions are typically small, irregularly shaped, and underrepresented in the dataset, even minor changes in regularization or optimization can lead to large fluctuations in the false negative rate of the fire class. This highlights the importance of tuning with a weighted objective function such as our $F(x)$, which explicitly penalizes missed fire detections while still maintaining strong performance across other segmentation classes. Similar patterns have been reported in studies of class‑imbalance effects, where minority classes exhibit disproportionate performance variability under different training regimes~\cite{buda2018imbalance}. These observations confirm the value of our weighted scoring approach for achieving reliable and fire-focused model performance.
% After selection, the best model was rebuilt using the chosen hyperparameters and compiled for deployment or further evaluation. The corresponding training history and evaluation metrics were retained for reproducibility.

%% file: sections/03-results.tex
\section{Results} \label{Results}

\subsection{Study 1: Statistical Comparison and Analysis of Different Augmentation Methods} \label{Results: Study01}

As outlined in Section \ref{Methodology: Eval_Aug_Methods}, \textbf{Study 1} investigates the performance impact of five data augmentation strategies: Non-Augmented (baseline), Rotation-by-15$^\circ$, Brightness Scaling, Contrast Scaling, and Standard Copy-Paste. The objective is to evaluate whether these methods, particularly Standard Copy-Paste, enhance model performance by introducing variability into the training data. The values of the relevant performance metrics for these five methods are reported in Table \ref{table:study01_comparison}. 

% study01 table
\begin{table}[!htb]
\centering
\caption{Study 1 performance metric [\%] outputs and weight-sum score across different augmentation methods.}
\label{table:study01_comparison}
\begin{tabular}{>{\centering\arraybackslash}p{5cm} c c c c}
\toprule
Method & Fire-FNR [\%] & Veg-IoU [\%] & Total-IoU [\%] & Score\\
\midrule
Non-Augmented & 16.16 & 60.83 & 45.26 & 0.7316\\
Rotation-by-15$^\circ$ & 11.74 & 64.24 & 55.46 & 0.7794\\
Brightness Scaling & 16.83 & 67.08 & 55.09 & 0.7558\\
Contrast Scaling & 11.68 & 67.62 & 55.98 & 0.7898\\
\textbf{Standard Copy-Paste} & \textbf{5.21} & \textbf{66.09} & \textbf{56.82} & \textbf{0.8259}\\
\bottomrule
\end{tabular}
\end{table}

Across the augmentation methods, a clear trend emerges: the Fire-FNR metric improves notably with most augmentation strategies. The baseline Non-Augmented method yields the highest Fire-FNR at 16.16\%, while the Standard Copy-Paste method achieves the lowest at 5.21\%, reflecting a substantial relative improvement of 67.8\% over the baseline. Contrast Scaling shows moderate improvement, reducing Fire-FNR to 11.68\%, while Rotation-by-15$^\circ$ achieves a similar value of 11.74\%. Brightness Scaling, however, does not improve Fire-FNR and records the highest among all augmented approaches at 16.83\%. Finally, the weighted score $F(x)$ provides a comprehensive summary by combining fire detection accuracy (quantified as 1-FNR) with segmentation quality across classes. Standard Copy-Paste yields the highest score at 0.8259, representing a 12.9\% relative improvement over the baseline score of 0.7316 and a 5.9\% increase over the score achieved by Rotation-by-15$^\circ$. Contrast Scaling follows with a score of 0.7898, further confirming its value as an adequate augmentation strategy. These results confirm that data augmentation, particularly the Standard Copy-Paste method, leads to notable improvements in key metrics, making it an effective strategy for enhancing model’s segmentation performance.

Next, we examine whether increasing the number of class-specific pixels directly correlates with improved segmentation performance. As shown in Table \ref{table:results_pixel_pct}, simply increasing the number of pixels in a particular class does not necessarily lead to better model performance. For instance, the Rotation-by-15$^\circ$ method results in 10.71\% fire pixels in the training data---more than the 9.83\% used in the Standard Copy-Paste method---yet its Fire-FNR is significantly higher (11.74\% vs. 5.21\%). This suggests that a higher quantity of class-specific pixels does not guarantee better segmentation performance. Despite having fewer fire-class pixels, the Standard Copy-Paste method outperforms the Rotation-by-15$^\circ$ approach. This can be attributed to the spatial variability introduced by pasting fire clusters at random locations across the training dataset, enhancing the model’s exposure to contextual variety. In contrast, the Rotation-by-15$^\circ$ method does generate additional fire pixels; however, simple rotation does not provide the meaningful spatial contextual diversity necessary for effective generalization. Thus, while rotation-based augmentation contributes to performance improvements by increasing pixel count, the spatial variability introduced by the Copy-Paste method has a more significant impact, even with fewer data points.

% visual image of study01
\begin{figure}[!htb]
    \centering
    \includegraphics[width=1.0\textwidth]{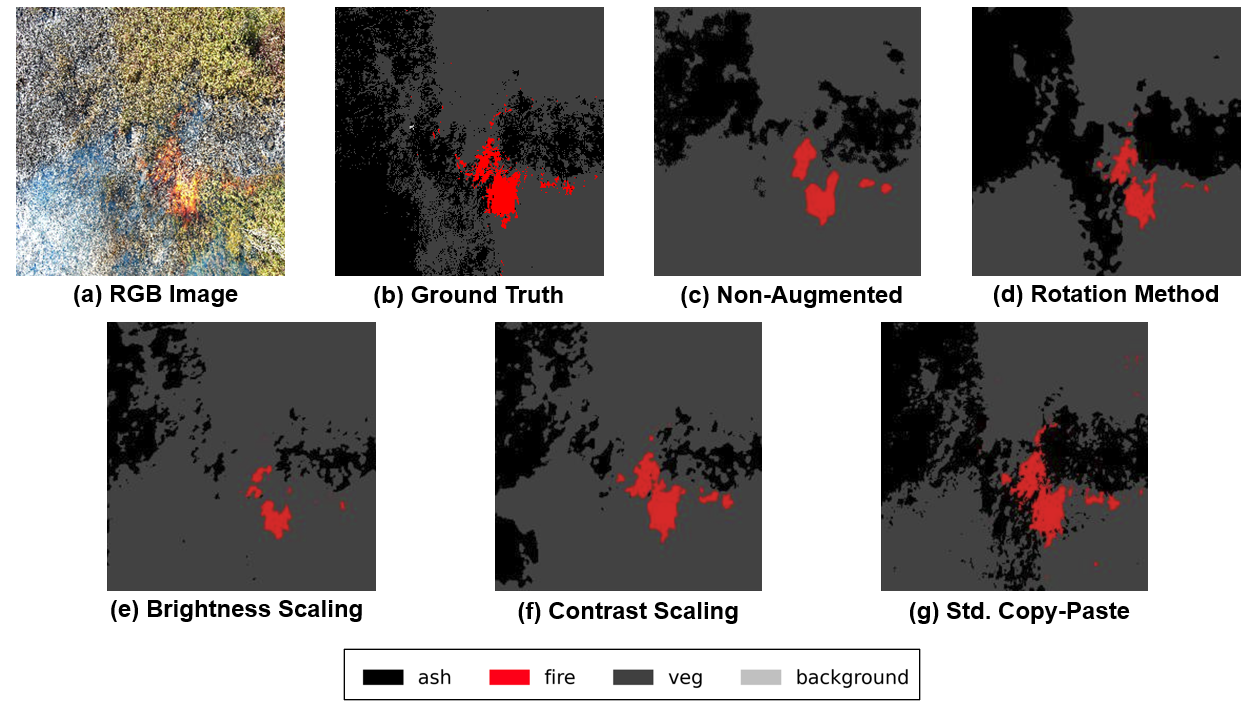}
    \caption{Visual Comparison of the Semantic Segmentation Results of \textbf{Study 1}.}
    \label{fig:visual_study01c}
\end{figure}

Fig.~\ref{fig:visual_study01c} presents a sample of visual inferences generated by the different augmentation methods under consideration. It confirms that the Standard Copy-Paste method outperforms all other augmentation strategies. When focusing on the fire class, we observe that fire cluster predictions become more detailed and accurate with the use of better augmentation techniques. The Standard Copy-Paste approach better captures the complex boundaries of the fire clusters, while the Non-augmented method, color-based scaling approaches, and the Rotation-by-15$^\circ$ method tend to overestimate the fire region, often producing rounded and overly generalized shapes. In addition, under-prediction is apparent in the Non-Augmented, Rotation, and Brightness methods, but this issue is notably corrected with the Standard Copy-Paste method. Although the Copy-Paste technique may occasionally lead to slight over-prediction in the fire class, false positives are preferable to false negatives in safety-critical contexts such as wildfire detection. Even when examining other classes, the segmentation results become more refined as augmentation improves. However, across all methods, the model sometimes struggles to distinguish between ash and vegetation, often predicting large areas of ash as vegetation. Despite these limitations, the Standard Copy-Paste method consistently delivers the best quantitative outcomes. %making it the prevailing strategy.

% study01 and study02 stats table 
\begin{table}[!htb]
\centering
\caption{Comparison of fire pixel percentages across different augmentation methods in Study 1 and Study 2, including total pixel counts for non-augmented and augmented datasets.}
\label{table:results_pixel_pct}
\begin{tabular}{>{\centering\arraybackslash}p{4cm} c||c c}
\toprule
\textbf{(Study 1)} Method & Fire [\%] & \textbf{(Study 2)} Erosion [\%] & Fire [\%] \\
\midrule
Non-Augmented         & 5.31  & 0\%   & 9.31 \\
Rotation-by-15$^\circ$ & 10.71 & 10\%  & 8.83 \\
Brightness Scaling     & 5.31  & 20\%  & 8.30 \\
Contrast Scaling       & 5.31  & 30\%  & 7.89 \\
Std. Copy-Paste        & 9.83  &       &      \\
\midrule
\multicolumn{2}{l}{Non-Aug. Data Pixel Total} & \multicolumn{2}{c}{524,288} \\
\multicolumn{2}{l}{Augmented Data Pixel Total} & \multicolumn{2}{c}{12,582,912} \\
\bottomrule
\end{tabular}
\end{table}

\subsection{Study 2: Performance Assessment and Analysis of CCPDA Method Using Different Erosion Values} \label{Results: Study02}

\textbf{Study 2} examines the effectiveness of the CCPDA approach with four erosion levels---0\%, 10\%, 20\%, and 30\%---as discussed in \ref{Methodology: Eval_Aug_Methods}. The goal is to mitigate human labeling errors at the boundaries of fire clusters, particularly where fire and vegetation pixels are visually similar due to overlapping RGB characteristics. From the results in Table \ref{table:study02_comparison}, applying 10\% erosion yields the best overall performance, achieving the lowest Fire-FNR (5.14\%) and the highest weighted score (0.8278). This indicates that modest erosion effectively removes ambiguous edge pixels that might otherwise introduce noise into the training data, providing empirical evidence that human labeling errors often occur near the boundaries of fire clusters due to the difficulty of distinguishing fire pixels from vegetation. Even with accurate human labeling, the model may still struggle to differentiate between the two classes because they share highly similar features. 

At 0\% erosion, Fire-FNR is slightly higher (7.18\%) and the weighted score is lower (0.8134), indicating that retaining all edge pixels may preserve labeling errors that degrade model accuracy. However, once the erosion percentage exceeds 10\%, the performance begins to decline. At 20\% erosion, the Fire-FNR increases to 6.29\%, and at 30\% erosion, it rises further to 7.60\%, with corresponding drops in the overall score to 0.8159 and 0.8093, respectively. This trend suggests that excessive erosion begins to remove correctly labeled fire pixels, resulting in a loss of valuable data and ultimately reducing model performance. Therefore, it is crucial to carefully explore and determine the optimal erosion value to balance minimizing labeling errors and preserving important information. Further refinement of erosion strategies remains a valuable direction for future work.

\begin{table}[!htb]
\centering
\caption{Study 2 performance metric [\%] outputs and weighted-sum score across different erosion percentages in the CCPDA method.}
\label{table:study02_comparison}
\begin{tabular}{>{\centering\arraybackslash}p{3cm} c c c c}
\toprule
Method & Fire-FNR [\%] & Veg-IoU [\%] & Total-IoU [\%] & Score \\
\midrule
Erosion-by-0\%  & 7.18  & 66.32 & 55.76 & 0.8134 \\
\textbf{Erosion-by-10\%} & \textbf{5.14} & \textbf{66.45} & \textbf{57.16} & \textbf{0.8278} \\
Erosion-by-20\% & 6.29  & 65.48 & 55.26 & 0.8159 \\
Erosion-by-30\% & 7.60  & 65.85 & 55.56 & 0.8093 \\
\bottomrule
\end{tabular}
\end{table}

%% file: sections/04-conclusion.tex
\section{Conclusion} \label{Conclusion}
In this paper, we explored a copy-paste methodology for data augmentation to improve multiclass segmentation performance in wildland fire image datasets. We first evaluated the efficacy of the Standard Copy-Paste augmentation method against other augmentation strategies. Our numerical findings and comparative analysis provide empirical evidence that the spatial variability introduced by the Standard Copy-Paste approach yields improved model performance compared to simpler techniques, such as rotation-based augmentation. Furthermore, to address the issue of mislabeling at the edges of fire clusters, we proposed a refinement of the standard method through the Centralized Copy-Paste Data Augmentation (CCPDA) approach. Experimental results confirm that applying moderate erosion (10\%) to copied fire segments can reduce edge-related labeling noise and improve segmentation accuracy. These findings suggest that targeted strategies such as CCPDA can play a critical role in enhancing model robustness for safety-critical applications, such as wildfire management and monitoring.  

In addition, this work created an original four-class wildfire dataset, BURN 1~\cite{BURN01}, comprising 20 RGB images with fully annotated labels. As discussed in Section~\ref{Introduction: DL Models and Data Constraints}, there is a scarcity of publicly available wildfire image datasets. In contrast to existing open-source datasets such as Flame~\cite{shamsoshoara2020dataset} and Flame 2~\cite{hopkins2022dataset}, which provide only binary labels (Fire or No Fire), the dataset developed in this study supports more detailed multiclass segmentation. It distinguishes between four classes---fire, vegetation, ash, and background---thereby offering valuable contributions to the wildfire research community.

Our future work will include evaluating the viability of the CCPDA method across various deep learning models, such as SegNet \cite{badrinarayanan2017SegNet} and DeepLab \cite{chen2018DeepLab}, to ensure its generalizability and sustained performance. Additionally, constraints will be implemented to avoid pasting fire clusters over existing fire regions or non-viable locations, such as parking lots, lakes, etc. A rules-based machine learning offers a starting point for this work \cite{raof2008colorthreshold}.